\documentclass{article}

\usepackage{microtype}
\usepackage{graphicx}
\usepackage{subcaption}
\usepackage{booktabs} 
\usepackage{amssymb}    
\usepackage[table]{xcolor}
\usepackage{booktabs}      
\usepackage{colortbl}
\usepackage{xcolor}
\usepackage{booktabs}
\usepackage{amssymb}    
\usepackage{tikz}
\usetikzlibrary{tikzmark}
\usepackage{colortbl}   
\usepackage{tikz} 
\usepackage{multirow}
\usetikzlibrary{tikzmark, arrows.meta, calc}
\usepackage{array}            
\usepackage{graphicx}
\newcommand{\cmarkNoCoT}{\cmark^{\text{\kern-2pt\scriptsize$-$}}}

\usetikzlibrary{tikzmark,calc}
\usepackage{amsmath}
\usepackage{amssymb}
\usepackage{mathtools}
\usepackage{amsthm}
\usepackage{tabularx}
\usepackage[most,skins,theorems]{tcolorbox}
\usepackage{tikz}
\usepackage{framed}
\usepackage{pifont}
\usepackage{hyperref}
\usepackage{framed}
\usepackage{ragged2e}

\usepackage[most]{tcolorbox}

\usepackage[preprint]{arxiv}

\usepackage[table]{xcolor}
\usepackage{amsmath}
\usepackage{amssymb}
\usepackage{mathtools}
\usepackage{amsthm}
\usepackage{tabularx}
\usepackage[most,skins,theorems]{tcolorbox}
\usepackage{xcolor} 
\usepackage{graphicx}
\usepackage{multirow} 
\usepackage[utf8]{inputenc}
\usepackage{booktabs}   
\usepackage{graphicx}   
\usepackage[table]{xcolor}
\usepackage{array}
\usepackage{makecell}
\usepackage{pifont}
\usepackage{xcolor} 
\usepackage[table]{xcolor} 
\usepackage{colortbl}
\usepackage{booktabs}
\usepackage{graphicx}
\usepackage{multirow}
\definecolor{myemerald}{rgb}{0.753, 0.898, 0.804}
\definecolor{mylightgreen}{rgb}{0.894, 0.933, 0.745}
\definecolor{myyellow}{rgb}{0.996, 0.972, 0.780}
\newcommand{\firstc}{\cellcolor{myemerald!100}}
\newcommand{\secondc}{\cellcolor{mylightgreen!100}} 
\newcommand{\thirdc}{\cellcolor{myyellow!100}}

\newcommand{\cmark}{\textcolor{green!60!black}{\ding{51}}}
\definecolor{backred}{rgb}{1,0.8,0.8}
\newcommand{\Gray}{\rowcolor[gray]{.9}}

\tcbset{
  aibox/.style={
    width=\linewidth,
    top=3pt,
    bottom=4pt,
    colback=blue!6!white,
    colframe=black,
    enhanced,
    center,
  }
}

\newtcolorbox{AIbox}[2][]{%
  aibox,
  #1,
  before upper={\textbf{#2}\par\smallskip},
}

\usepackage[capitalize,noabbrev]{cleveref}

\theoremstyle{plain}

\theoremstyle{definition}

\theoremstyle{remark}

\usepackage[textsize=tiny]{todonotes}

\definecolor{wdcolor}{RGB}{128, 0, 255}

\definecolor{wdqcolor}{RGB}{255, 0, 0}

\icmltitlerunning{}

\begin{document}

\twocolumn[
  \icmltitle{GeoEyes: On-Demand Visual Focusing for Evidence-Grounded Understanding of Ultra-High-Resolution Remote Sensing Imagery}



  \icmlsetsymbol{equal}{*}

  \begin{icmlauthorlist}
    \icmlauthor{Fengxiang Wang}{nudt}
    \icmlauthor{Mingshuo Chen}{bupt}
    \icmlauthor{Yueying Li}{nudt}
    \icmlauthor{Yajie Yang}{ucas}
    \icmlauthor{Yifan Zhang}{cas}\textsuperscript{*}
    \icmlauthor{Long Lan}{nudt}\textsuperscript{*} \\
    \icmlauthor{Xue Yang}{sjtu}
    \icmlauthor{Hongda Sun}{gaoling}\textsuperscript{*}
    \icmlauthor{Yulin Wang}{thu}
    \icmlauthor{Di Wang}{whu}
    \icmlauthor{Jun Song}{}
     \icmlauthor{Jing Zhang}{whu}
    \icmlauthor{Bo Du}{whu}
  \end{icmlauthorlist}
  \icmlaffiliation{nudt}{National University of Defense Technology, China}
  \icmlaffiliation{bupt}{Beijing University of Posts and Telecommunications, China}
  \icmlaffiliation{ucas}{University of the Chinese Academy of Sciences, China}
  \icmlaffiliation{sjtu}{Shanghai Jiao Tong University, China}
  \icmlaffiliation{whu}{Wuhan University, China}
  \icmlaffiliation{cas}{Chinese Academy of Science, China}
  \icmlaffiliation{thu}{Tsinghua University, China}
  \icmlaffiliation{gaoling}{Renmin University of China, China}
  \icmlcorrespondingauthor{Yifan Zhang}{}
  \icmlcorrespondingauthor{Long Lan}{}
  \icmlcorrespondingauthor{Hongda Sun}{}


  \vskip 0.05in
    \centerline{
        GitHub Repo: \url{https://github.com/nanocm/GeoEyes}
    }
  \vskip 0.3in
]



\printAffiliationsAndNotice{}  

\begin{abstract}

The “thinking-with-images” paradigm enables multimodal large language models (MLLMs) to actively explore visual scenes via zoom-in tools. This is essential for ultra-high-resolution (UHR) remote sensing VQA, where task-relevant cues are sparse and tiny. However, we observe a consistent failure mode in existing zoom-enabled MLLMs: Tool Usage Homogenization, where tool calls collapse into task-agnostic patterns, limiting effective evidence acquisition. To address this, we propose GeoEyes, a staged training framework consisting of (1) a cold-start SFT dataset, UHR Chain-of-Zoom (UHR-CoZ), which covers diverse zooming regimes, and (2) an agentic reinforcement learning method, AdaZoom-GRPO, that explicitly rewards evidence gain and answer improvement during zoom interactions. The resulting model learns on-demand zooming with proper stopping behavior and achieves substantial improvements on UHR remote sensing benchmarks, with 54.23\% accuracy on XLRS-Bench.

\end{abstract}

\section{Introduction}


Advances in Earth science increasingly depend on effective collection, processing, and interpretation of remote sensing (RS) data. 
Ultra-high-resolution (UHR) satellite imagery, in particular, captures fine-grained land object structures that are critical for understanding complex Earth surface patterns. 
However, in UHR RS scenarios, task-relevant cues, such as small objects or subtle structural patterns, often occupy only tiny fractions of the full scene~\cite{geollava}, making effective information extraction inherently challenging.
\begin{figure}[htbp]
\centering
\vspace{-1mm}
\includegraphics[width=0.99\columnwidth]{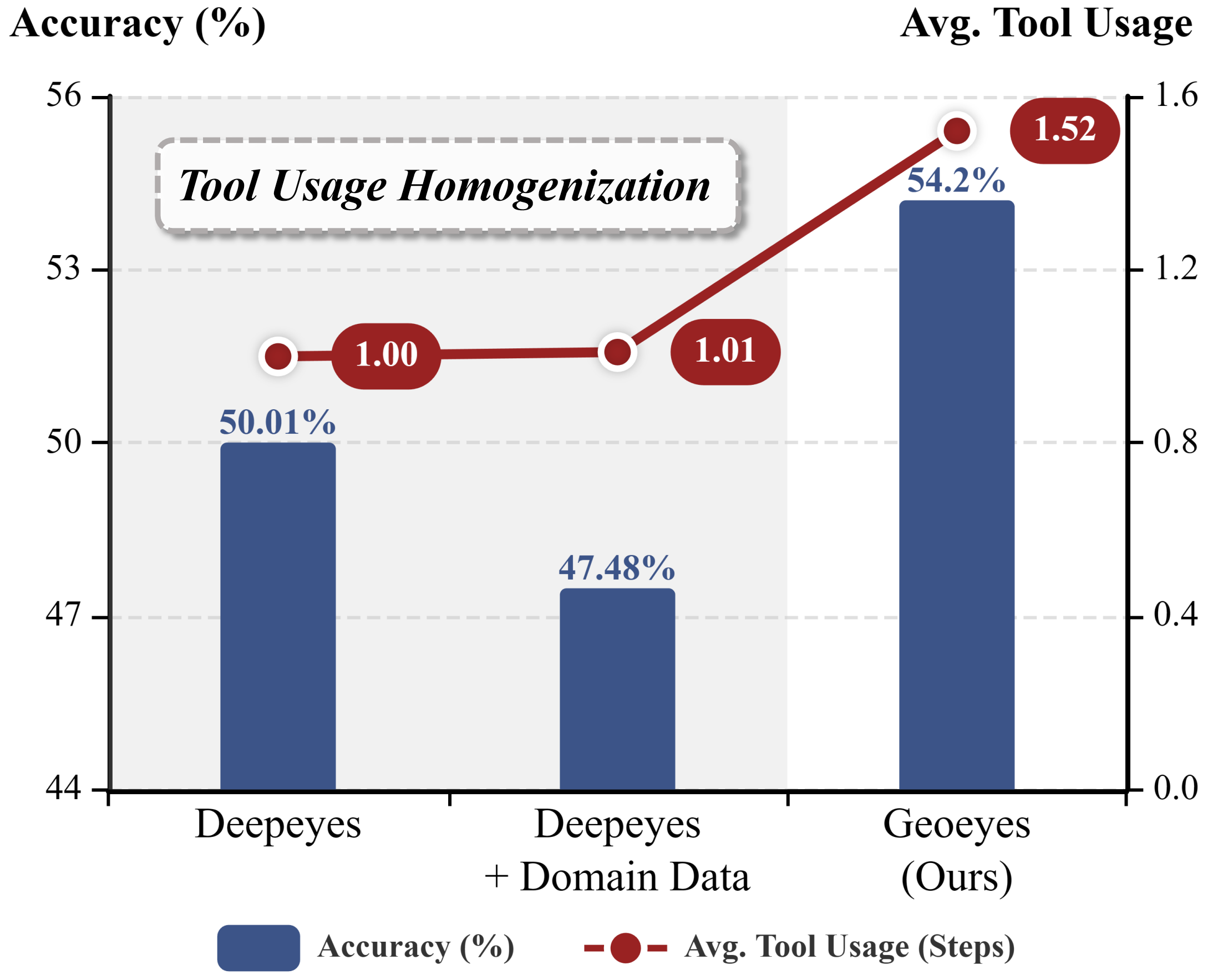}
\caption{\textbf{Illustration of the ``Tool Usage Homogenization'' phenomenon in UHR RS benchamrk.} \textit{Domain Data} means SuperRS-VQA~\cite{geollava}. \textit{Tool Usage Homogenization}: a collapse to a near-constant one-call tool pattern across samples. \textit{Avg. Tool Usage} means tool-call depth over samples that invoke the tool. Our GeoEyes  triggers the tool on 68.44\% of evaluation samples, compared with 100\% for Deepeyes and its domain-augmented variant.
}
\vspace{-1mm}
\label{fig:gap1}
\end{figure}
\begin{figure*}[t]

\centering
\includegraphics[width=0.99\textwidth]{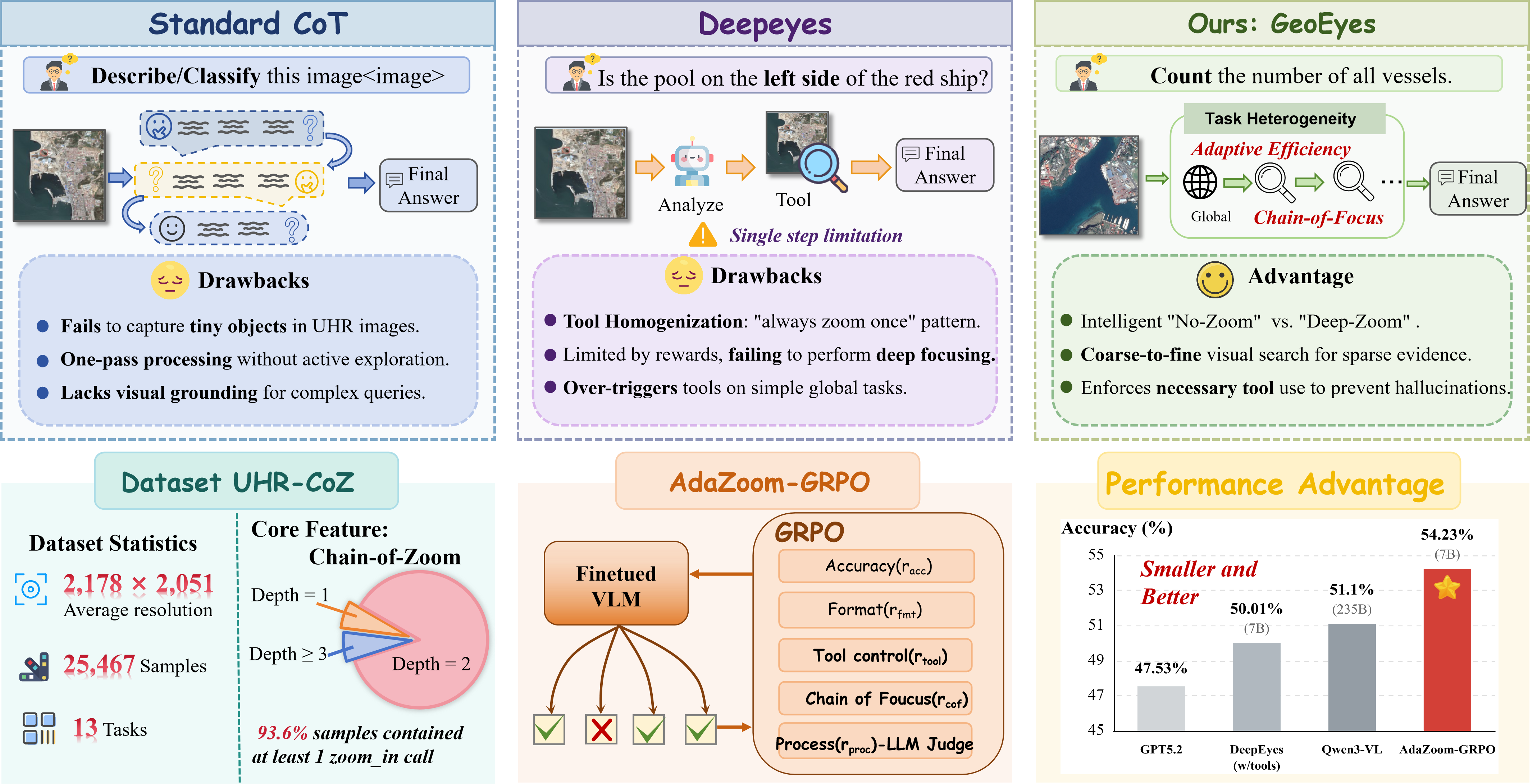}
\caption{GeoEyes enables task-adaptive tool use for UHR remote-sensing reasoning, from tool-free inference to multi-round progressive zooming. Specially, we introduce UHR-CoZ, an interleaved image–text CoT dataset, and AdaZoom-GRPO, a tailored RL method that trains GeoEyes to use tools for evidence gain. 
GeoEyes significantly outperforms state-of-the-art closed- and open-source baselines.
}
\label{fig:overview}
\vspace{-2mm}
\end{figure*}
Recently, the ``thinking-with-images'' paradigm has emerged as a promising approach for addressing this challenge by enabling active visual exploration. 
By interleaving textual reasoning with evidence acquired from dynamically selected image regions, visually grounded reasoning models can zoom into relevant areas during inference to extract fine-grained cues~\cite{deepeyes}. 
In UHR remote sensing tasks, effective reasoning relies on tight coupling between spatial evidence acquisition and reasoning, while the need for tool usage varies sharply across subtasks.
This raises a critical question: can models learn task-adaptive zoom-in policies that selectively abstain when unnecessary and progressively focus when needed?
In this work, we investigate this question through a staged empirical study on a UHR remote sensing benchmark within a representative thinking-with-images framework.

Our motivating experiments in Fig.~\ref{fig:gap1} reveal a consistent pattern of homogeneous tool usage on XLRS-Bench~\cite{xlrsbench}, a widely used UHR remote sensing benchmark with extremely high image resolution. Specifically, DeepEyes \cite{deepeyes}, used as the zoom-in agentic reinforcement learning (RL) baseline, achieves 50.01 accuracy but invokes the tool for every question. Incorporating the UHR remote sensing dataset SuperRS-VQA \cite{geollava} during RL training further reduces accuracy to 47.48, while tool usage remains fully saturated. These results indicate that once tool usage collapses into a uniform pattern, domain-data–only RL fails to induce task-adaptive tool policies or improve generalization performance.

We attribute this \emph{Tool Usage Homogenization} phenomenon to two UHR-specific challenges: \textbf{task heterogeneity} and \textbf{low effective evidence density}. 
(1) \textbf{Task heterogeneity} refers to the sharp variation in tool requirements across subtasks: some questions are solvable from a global view, where zooming adds noise and cost, while others require multiple rounds of focused inspection, making a single tool call insufficient. A uniform calling strategy therefore leads to both over-triggering on tool-irrelevant questions and under-exploration on tool-critical ones. 
(2) \textbf{Low effective evidence density} further exacerbates this issue. For a given question, most image regions provide negligible information gain, such that many tasks require progressive, multi-step focusing. Under supervision only from final answers, tool policies tend to collapse into stable but inefficient templates, consistent with the near-saturated single-call behavior observed in Fig. \ref{fig:gap1}. Together, these observations expose a critical issue in existing zoom-enabled MLLMs, which we aim to address in this work:


\begin{tcolorbox}[top=1pt, bottom=1pt, left=1pt, right=1pt]
\justifying
\textit{How can tool-augmented MLLMs learn task-adaptive zoom-in policies with selective activation and proper stopping in UHR remote sensing?}
\end{tcolorbox}


To address this challenge, we propose GeoEyes, which enables on-demand zooming for UHR remote sensing VQA. We first construct an interleaved reasoning-chain dataset, UHR Chain-of-Zoom (UHR-CoZ), by transforming HighRS-VQA~\cite{geollava}, explicitly covering diverse zoom-in regimes including no tool use, single-call, and multi-round progressive focusing.
Building on this dataset, we perform supervised fine-tuning to learn task-conditioned tool usage and stopping behavior. We then introduce AdaZoom-GRPO, an agentic RL stage with a novel reward that directly encourages evidence gain and answer improvement during zoom-in interactions.
As a result, GeoEyes enables adaptive zooming and effective multi-step focusing, yielding substantial performance gains on UHR remote sensing VQA tasks.

In summary, we make three contributions:


\begin{enumerate}
    \item We identify a tool-policy collapse phenomenon in UHR remote sensing scenarios, characterized by saturated tool usage with near-constant single-call behavior, and attribute it to task heterogeneity and low effective evidence density.
    \item We construct UHR Chain-of-Zoom (UHR-CoZ), an interleaved image-text chain-of-thought dataset for cold-start supervised fine-tuning, which explicitly covers no-tool, single-call, and multi-step progressive focusing regimes.
    \item We develop GeoEyes, a remote sensing MLLM tailored for UHR scenes, which learns adaptive zoom-in visual exploration with proper stopping via cold-start SFT and the proposed AdaZoom-GRPO training strategy. Extensive experiments on representative UHR RS benchmarks demonstrate its clear advantages over existing open- and closed-source MLLMs.
\end{enumerate}

\section{Related Work}
\textbf{MLLMs in Remote Sensing.}
General-purpose Multimodal Large Language Models (MLLMs) \cite{llava, interns1} have demonstrated exceptional visual understanding capabilities, inspiring the rapid development of specialized models within the RS domain. Early research primarily focused on aligning visual encoders with LLMs; such as RSGPT \cite{rsgpt}, SkyEyeGPT \cite{skyeyegpt}, GeoChat \cite{geochat}, EarthGPT \cite{earthgpt}. Subsequent studies have further refined these alignment mechanisms, including VHM \cite{VHM}, RS-CapRet \cite{RScapret}, EarthMarker \cite{Earthmarker} , LHRS-Bot-Nova \cite{LHRSBotNova}, RSUniVLM \cite{RSUniVLM}, EarthMind \cite{EarthMind}, EarthDial\cite{Earthdial},  RingMoGPT \cite{Ringmogpt}, and EarthVL \cite{EarthVL}.
However, in UHR RS scenarios, these models struggle to accurately locate task-relevant fine-grained regions within vast pixel spaces. Existing research has proposed various methodologies: Tokens Pruning methods in MLLMs (GeoLLaVA-8K \cite{geollava}, ImageRAG \cite{imagerag}, and RFM \cite{lrsvqa}), Tool-calling (ZoomEarth~\cite{LRSGRO}, VICoT-Agent \cite{vicot}).
Tokens Purning methods lack the ability to actively focus on specific regions based on task requirements.
While existing tool-calling methods, such as ZoomEarth, are mostly limited to single-turn zoom-in actions.
Such straightforward training leads models into Tool Usage Homogenization.

\textbf{Thinking with Images Framework.}
To overcome the limitations of text-only reasoning, the \textit{Thinking with Images} paradigm \cite{thinkwithimage} extends the Chain-of-Thought (CoT) framework by allowing models to interleave visual perception steps with textual reasoning steps \cite{deepeyes, vicot}.
In the general domain, agentic frameworks have demonstrated that actively querying visual tools can significantly enhance performance in complex reasoning tasks. Representative works include DeepEyes \cite{deepeyes}, DeepEyesV2 \cite{deepeyesv2}, SenseNova-MARS \cite{mars}, and Deep But Reliable \cite{deepreliable}.
Furthermore, recent approaches have explored various visual reasoning aids, such as Visual Sketchpad \cite{visualsketchpad}, Chain-of-Focus \cite{cof}, Pixel Reasoner \cite{pixelreasoner}, and DyFo \cite{dyfo}.
Specific strategies have also been proposed to enhance precision, including Thinking with Blueprints \cite{blueprints}, Thinking with Bounding Boxes \cite{boundingboxes}, Thinking with Map \cite{thinkingmap}, SpaceTools \cite{spacetools}, and ZoomEye \cite{zoomeye}.

\textbf{Interleaved Reasoning Datasets.}
In the general domain, datasets \cite{cof,deepeyes,zebra,visresason} have been proposed to bridge the reasoning gap, with VisReason \cite{visresason} providing large-scale benchmarks for visual reasoning. These datasets provide not only questions and answers but also document the complete trajectory—from browsing the global image to locking onto local details.
Remote sensing VQA datasets, such as RSVQA \cite{rsvqa}, LRSVQA \cite{lrsvqa}, and EarthVQA \cite{earthvqa}, primarily consist of simple QA pairs.
LRS-GRO-train~\cite{LRSGRO} dataset only provided 3,500 samples which has one turn zoom-in action for training.


\section{Dataset}
To address task heterogeneity and low effective evidence density in the SFT cold start, we build UHR-CoZ, an interleaved image–text chain-of-thought dataset.

\begin{table}[t]
    \centering
    \scriptsize
    \setlength{\tabcolsep}{3.5pt}
    \renewcommand{\arraystretch}{1.2}
    \definecolor{RowHi}{RGB}{235,244,255}
    \definecolor{RowAlt}{RGB}{248,249,250}
    \definecolor{ColHeader}{RGB}{245,247,250}
    \definecolor{GroupHeader}{RGB}{245,247,250}
    \definecolor{GenBlue}{RGB}{46,116,181}
    \definecolor{RSOrange}{RGB}{217,105,0}
    \definecolor{TxtPurple}{RGB}{124,77,255}
    \definecolor{diffgreen}{RGB}{50,160,65}
    \definecolor{diffred}{RGB}{220,50,50}
    
    \newcommand{\xmark}{\textcolor{gray!40}{\texttimes}}
    \newcommand{\genmark}{\raisebox{0.15ex}{\textcolor{GenBlue}{$\large\triangle$}}}
    \newcommand{\rsmark}{\raisebox{0.10ex}{\textcolor{RSOrange}{$\large\blacksquare$}}}
    \newcommand{\bothmark}{\genmark\hspace{0.1em}{\footnotesize+}\hspace{0.1em}\rsmark}
    \newcommand{\txtmark}{\raisebox{0.10ex}{\textcolor{TxtPurple}{$\large\blacklozenge$}}}
    \newcommand{\sftrsmark}{\raisebox{0.10ex}{\textcolor{RSOrange}{$\large\blacksquare$}}}
    \newcommand{\sftbothmark}{\txtmark\hspace{0.1em}{\footnotesize+}\hspace{0.1em}\sftrsmark}

    \resizebox{0.95\columnwidth}{!}{%
    \begin{tabular}{@{}l c c c r @{}}
    \toprule
    \rowcolor{ColHeader}
    \textbf{Base Model} & \textbf{SFT Data} & \textbf{RLVR Method} & \textbf{RLVR Data} & \textbf{Accuracy} \\
    \midrule
    \rowcolor{GroupHeader}
    QwenVL2.5 & \xmark & GRPO w/ tools & \genmark & 50.01 \tikzmark{group_gen} \\ 
    \rowcolor{RowAlt}
     \rowcolor{RowHi}
    QwenVL2.5 & \txtmark & GRPO w/ tools & \bothmark & \textbf{47.86}\textcolor{red}{$\downarrow $ } \tikzmark{group_dom} \\
    \bottomrule
    \end{tabular}%
    }
    \caption{\textbf{Pilot Experiments of Dataset. } GRPO w/ tools means the Deepeyes~\cite{deepeyes} method.
Data icons: \xmark{} none, \txtmark{} General data (MM-Adaptive-CoF), \sftrsmark{} Domain data (SuperRS-VQA), \genmark{} General data (DeepEyes-47K). }
    \label{tab:data_pilot}
    \vspace{-8mm}
\end{table}
\textbf{Pilot Experiments.} Since lacking the RS training data with multi-round tool-augmented CoT trajectories, we explore the general-domain MM-Adaptive-CoF dataset~\cite{cof}, which contains multi-step tool-use traces (2,201 samples with one call and 1,454 with two). We hypothesize that SFT on MM-Adaptive-CoF can teach adaptive tool use, after which RL on SuperRS-VQA can specialize it for remote sensing. 
However, as shown in Table~\ref{tab:data_pilot}, this general-domain multi-tool CoT supervision provides no gains. We argue this is because low effective evidence density is substantially more severe in satellite imagery, so general-domain tool traces alone cannot induce the multi-round tool-use behavior needed for UHR remote sensing VQA. Therefore, it is necessary to construct remote sensing UHR SFT training data that includes multi-round tool use.

\begin{figure*}[t]
  \centering
  \includegraphics[width=\textwidth]{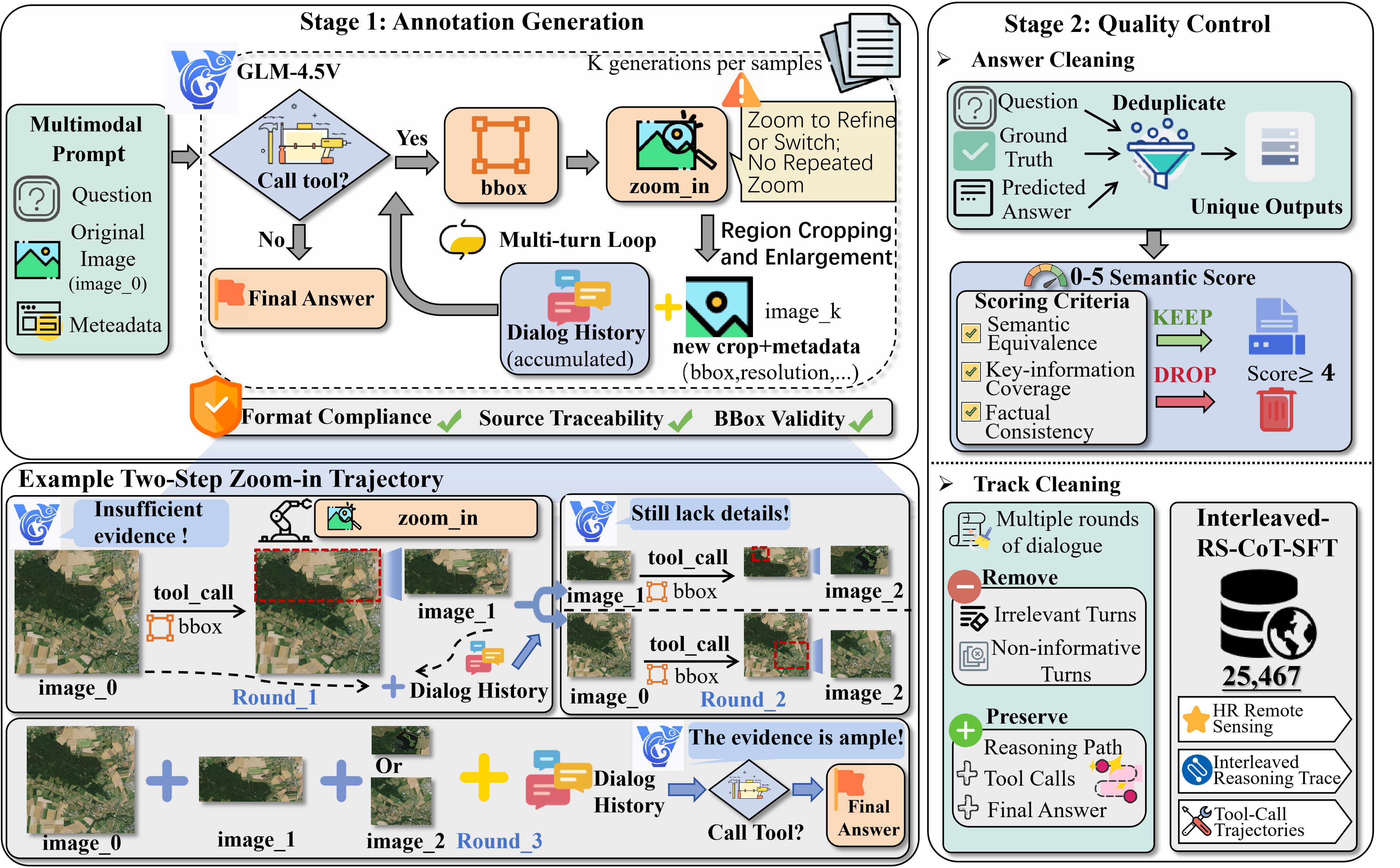}
  \caption{\textbf{Automated data construction pipeline.} Stage 1 performs UHR-CoF annotation generation with multi-round \texttt{zoom\_in}. Stage 2 applies quality control, including answer cleaning and trajectory cleaning. An example two-step \texttt{zoom\_in} trajectory is shown to illustrate the multi-round generation process.}
  \label{fig:data_pipeline}
  \vspace{-4mm}
\end{figure*}

\noindent \textbf{Data Sources.} 
UHR-CoZ is derived from HighRS-VQA~\cite{geollava}, which features 2K-resolution imagery with low effective evidence density, as relevant cues are sparsely distributed across large scenes.

\noindent \textbf{Data Construction Pipeline.} 
To efficiently construct a cold-start dataset of interleaved image–text reasoning chains, we develop an automated, agent-orchestrated pipeline centered on GLM-4.5V~\cite{GLM45v} as shown in Figure~\ref{fig:data_pipeline}.
It runs in two stages, annotation generation and quality control, converting HR RS-VQA inputs into SFT samples with reasoning text and adaptive tool-call trajectories.

\noindent (1) \textbf{Annotation Generation for UHR-CoZ.} 
The main challenge in annotation generation is not answering questions, but reliably localizing evidence in high-resolution images and annotating the visual evidence. 
In HighRS-VQA-like settings, targets are small and cues are sparse, so global views with the general MLLMs often fall short. 
Tool decisions therefore require iterative refinement, including whether to zoom, where to zoom, and how many times.
We address this challenge with a agent pipeline, where an agent orchestrates the model and a \texttt{zoom\_in} tool to generate end-to-end reasoning trajectories. 
Representative multi-round trajectories and sample demonstrations are provided in Appendix~\ref{app:uhrcof_examples}.

\noindent (2) \textbf{Quality Control.} 
We apply two-step quality control to the generated results: answer cleaning and reasoning trajectory cleaning.
For answer cleaning, the agent aggregates the question, reference answer, and predicted answer, then uses a dedicated scoring prompt to assign a 0--5 semantic-consistency score based on semantic equivalence, key-information coverage, and factual consistency.
We deduplicate multiple generations per question and retain only samples with scores $\geq$ 4. 
For reasoning trajectory cleaning, we normalize multi-turn reasoning trajectories by removing irrelevant or non-informative turns, while preserving the reasoning path, valid tool calls, and the final answer. The output is structured SFT samples with essential metadata.

\begin{table}[!t]
\centering
\caption{\textbf{Main statistics of the UHR-CoZ dataset.}
\textbf{Note:} Question length and reasoning length are measured in tokens. Tool-used samples denote the fraction of samples with at least one \texttt{zoom\_in} call. Average tool calls per sample averages over all samples, counting zero for tool-free samples.}
\label{tab:dataset_stats}
\small
\setlength{\tabcolsep}{6pt}
\begin{tabularx}{\columnwidth}{@{}>{\raggedright\arraybackslash}X r@{}}
\toprule
\textbf{Statistics} & \textbf{Value} \\
\midrule
Total samples & 25,467 \\
Average image resolution & 2,178\(\times\)2,051 \\ 
Average question length & 103.2 \\
Average reasoning length & 157.8 \\
Tool-used samples & 93.6\% \\
Average tool calls per sample & 1.1 \\
\midrule
Zoom-in-chain depth \(D\) distribution & \\
\quad \(D=1\) (no zoom-in) & 6.4\% \\
\quad \(D=2\) (one zoom-in) & 86.7\% \\
\quad \(D\geq 3\) (deeper zoom-in) & 6.9\% \\

\bottomrule
\end{tabularx}
\vspace{-6mm}
\end{table}

\noindent \textbf{Statistics.}
To the best of our knowledge, UHR-CoZ is the largest cold-start dataset for HR RS that systematically annotates interleaved multi-turn tool-use reasoning trajectories. 
It covers diverse remote sensing tasks in real-world settings, including Object Classification, Object Color, Object Special Relationship, Overall Counting, Route planning, Anomaly Detection, Environmental Condition Reasoning and so on.
We further define zoom-in-chain depth to quantify tool-augmented reasoning. Depth = 1 indicates no zoom-in; Depth = 2 indicates at least one zoom-in from global to local;  and Depth $\geq$ 3 indicates deeper progressive zoom-in. This metric highlights heterogeneity in tool-call requirements across samples. Key dataset statistics are reported in Table~\ref{tab:dataset_stats}.

\section{Method}
This section details the improved framework based on Zoom-in Agentic RL method. We first review the Zoom-in Agentic RL mechanism and analyze the root causes of "Tool Usage Homogenization" in UHR RS scenarios. 
Subsequently, we detail the our proposed method, covering the full process from cold-start SFT and new reward for RL optimization.

\subsection{Preliminaries: Zoom-in Agentic RL Paradigm}
We use the DeepEyes~\cite{deepeyes} as the baseline of Zoom-in Agentic RL method. 
It introduces a tool call mechanism into visual reasoning.
Specifically, given an image $I$ and a question $q$, the model generates a trajectory interleaved with textual Chain-of-Thought and visual actions (Zoom-in action).
The model autonomously decides whether to continue reasoning or invoke tools to acquire high-resolution details based on the current state. 
This process is driven by end-to-end reinforcement learning, ultimately rewarded based on the correctness of the answer.

\textbf{Challenges in UHR Scenarios.} While this paradigm performs well in general scenarios, it suffers from severe tool usage policy homogenization in UHR RS benchmarks (e.g., XLRS-Bench), mechanically executing tool calls for all samples. This fails to adapt to two essential UHR RS characteristics: 

\textbf{(1) Task Heterogeneity}, where task difficulty varies significantly, global tasks (e.g., global classification) require no tools, whereas fine-grained tasks (e.g., vehicle counting) are tool-reliant. DeepEyes' original reward encourages zoom-in action for reward, causing computational waste.

\textbf{(2) Low Effective Evidence Density}, where effective information in $8,500 \times 8,500$ images is extremely sparse. Relying solely on sparse outcome rewards fails to guide the DeepEyes in multi-step searching, leading to degeneration into local optima.

\subsection{Our Approach}
We propose a two-stage training scheme: a Cold-Start SFT on a constructed UHR-CoZ dataset to initialize basic domain tool capabilities, followed by Reinforcement Learning with a new reconstructed reward function to address UHR RS specific difficulties. 
\begin{figure*}[htbp]
\centering
\includegraphics[width=0.99\textwidth]{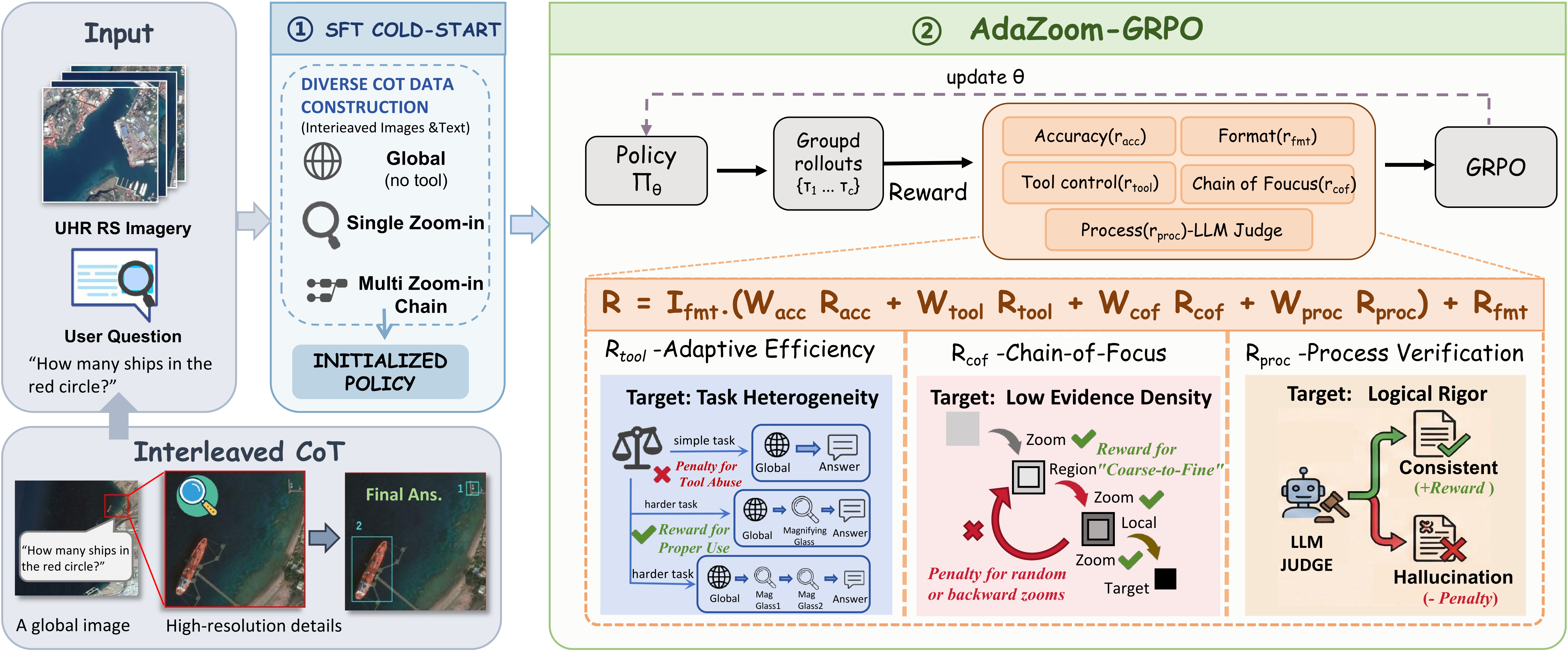}
\caption{\textbf{Overview of our method.} We first perform cold-start SFT on UHR-CoZ, then apply AdaZoom-GRPO for RL. AdaZoom-GRPO mainly includes an Adaptive Efficiency reward for task heterogeneity, a Chain-of-Focus reward for low evidence density, and a Process Verification reward to enforce logical rigor.
}
\label{fig:method}
\vspace{-6mm}
\end{figure*}

\subsubsection{Cold-start SFT}
To mitigate exploration challenges in the massive search space, we utilize the UHR-CoZ dataset for cold-start SFT.
Crucially, UHR-CoZ \textbf{encompasses} three distinct reasoning regimes: (1) Tool-free trajectories for global tasks; (2) Single-step zoom interactions for medium-scale targets; and (3) Multi-step progressive focusing paths for tiny objects.
Supervised training on these diverse demonstrations initializes the policy $\pi_{SFT}$ with fundamental visual planning capabilities and task difficulty awareness, establishing a stable starting point for the subsequent RL stage.

\subsubsection{AdaZoom-GRPO} 
In the RL phase, we proposed the reward function to teach correct tool usage and deep reasoning. The total reward $R$ is a format-gated weighted combination:
\begin{equation}
\begin{aligned}
    R = &~ \mathbb{I}_{fmt} \cdot ( w_{acc}R_{acc} + w_{tool}R_{tool} \\
        & + w_{cof}R_{cof} + w_{proc}R_{proc} ) + R_{fmt}
\end{aligned}
\end{equation}
where $w_{acc}, w_{tool}, w_{cof},$ and $w_{proc}$ are the weight coefficients assigned to their respective reward terms. The indicator function $\mathbb{I}_{fmt}$ acts as a logical gate, ensuring that task-specific rewards are only granted when the output strictly adheres to the our defined conversation protocol.
Beyond standard accuracy ($R_{acc}$) and format ($R_{fmt}$) rewards, we introduce three core components:

\textbf{Addressing Task Heterogeneity: Adaptive Efficiency Reward ($R_{tool}$).}
To balance tool usage across heterogeneous tasks, we introduce a dynamic efficiency mechanism that adapts to both category-level complexity and instance-level difficulty.
The reward is formulated as $R_{tool} =  P_{\alpha} \cdot \exp\left(-\gamma \cdot \Delta N\right)$, where $N_{step}$ represents the \textbf{actual number of tool execution steps}, and the excess step count is defined as $\Delta N = \max(0, N_{step} - N_{base}(C))$.
This mechanism operates through two layers of control:
\begin{itemize}
    \item \textbf{Category-Specific Step Allowance ($N_{base}^{(C)}$):} This term establishes a "free" tool-use quota based on the intrinsic difficulty of category $C$ (e.g., $N_{base}=1$ for global tasks, $N_{base}=2$ or higher for tiny object detection). The exponential decay penalty only activates when the agent's steps exceed this predefined allowance, tolerating necessary exploration for hard tasks while strictly limiting it for simple ones.
    \item \textbf{Instance-Level Difficulty Modulation ($P_{\alpha}$):} Simultaneously, we introduce a dynamic scalar $P_{\alpha} = 1 - p(y|x)_{base}$ to quantify sample difficulty.
    For easy samples ($P_{\alpha} \approx 0$) where the problem is solvable via the base model's intrinsic perception or text-only reasoning, the efficiency reward is suppressed, discouraging the agent from invoking further tools to exploit rewards, which would incur unnecessary computation.
    Conversely, for hard samples ($P_{\alpha} \approx 1$), the reward magnitude is amplified, incentivizing the agent to actively utilize tools to solve the problem.
\end{itemize}

\textbf{Addressing Low Evidence Density: Chain-of-Focus Reward ($R_{cof}$).}
To mitigate the search inefficiency caused by sparse visual cues, we design $R_{cof}$ to enforce a structured "Coarse-to-Fine" trajectory.
Instead of allowing stochastic exploration, we evaluate the geometric containment between consecutive view windows $b_t$ and $b_{t+1}$:
\begin{equation}
r_t^{cof} = \begin{cases} 
  +\beta_{zoom}, & \begin{aligned}[t] 
      &\text{if } b_{t+1} \subset b_t \text{ and } \\[-2pt] 
      &\text{Area}(b_{t+1}) < \text{Area}(b_t) 
  \end{aligned} \\
  0, & \text{if } b_t \subset b_{t+1} \text{ (Backtrack)} \\
  -\beta_{drift}, & \text{otherwise}
\end{cases}
\end{equation}
where $\text{Area}(\cdot)$ denotes the area of the bounding box, and $\beta_{zoom}, \beta_{drift} > 0$ are hyperparameters regulating the reward magnitudes. Note that this equation presents a simplified version for conceptual clarity; in implementation, we incorporate additional logic to handle boundary conditions.
This formulation incentivizes effective Zoom-in actions, where the agent strictly narrows its scope to isolate details.
Crucially, we assign a neutral reward to \textbf{Backtracking} (Context Expansion, $b_t \subset b_{t+1}$).
This provides a "safe harbor" for error recovery, allowing the agent to actively correct localization mistakes by reverting to a broader view without penalty, while penalizing disjoint drift to prevent aimless browsing.

\begin{table*}[htbp]
\footnotesize
\centering
\caption{\textbf{Experimental results on the perception and reasoning dimensions on XLRS-Bench.} 
The sub-tasks include \textbf{Perception}: Overall/Regional Counting (OC/RC), Overall/Regional Land Use Classification (OLUC/RLUC), Object Classification (OCC), Object Color (OCL), Object Motion State (OMS), and Object Spatial Relationship (OSR); and \textbf{Reasoning}: Anomaly Detection and Interpretation (AD), Environmental Condition Reasoning (ECR), Route Planning (RP), Regional Counting with Change Detection (RCCD), and Counting with Complex Reasoning (CCR). 
`Avg.' represents the macro average accuracy across sub-tasks. 
Rankings are calculated column-wise: \colorbox{myemerald}{green} (1st, bold), \colorbox{mylightgreen}{light green} (2nd), and \colorbox{myyellow}{yellow} (3rd). Tie scores are treated identically.}
\label{tab:vqa}

\resizebox{\textwidth}{!}{%
\begin{tabular}{l|cccccccc|ccccc|c}
\toprule
\rowcolor[gray]{.9} \textbf{Method} & \multicolumn{8}{c|}{\textbf{Perception}} & \multicolumn{5}{c|}{\textbf{Reasoning}} & \textbf{Avg.} \\
\midrule
\rowcolor[gray]{.9} \textbf{Sub-tasks (L-3 Capability)} & \textbf{OC} & \textbf{RC} & \textbf{OLUC} & \textbf{RLUC} & \textbf{OCC} & \textbf{OCL} & \textbf{OMS} & \textbf{OSR} & \textbf{AD} & \textbf{ECR} & \textbf{RP} & \textbf{RCCD} & \textbf{CCR} & \\
\midrule
\multicolumn{15}{l}{\textit{\textcolor{gray}{Remote Sensing MLLMs}}} \\
GeoChat~\cite{geochat} & 16.7 & 29.0 & 2.0 & 23.0 & 21.1 & 16.8 & 35.0 & 24.2 & 33.0 & 43.0 & 10.0 & -- & 21.0 & 22.9 \\
ZoomEarth~\cite{LRSGRO} & -- & -- & -- & -- & -- & -- & -- & -- & -- & -- & -- & -- & -- & 40.2 \\
GeoLLaVA-8K~\cite{geollava} & 26.7 & 38.0 & \firstc \textbf{49.0} & 69.0 & 41.6 & 31.6 & \thirdc 65.0 & \thirdc 35.0 & 67.0 & 78.0 & \firstc \textbf{66.0} & \firstc \textbf{50.0} & \firstc \textbf{52.0} & \thirdc 51.5 \\
\midrule
\multicolumn{15}{l}{\textit{\textcolor{gray}{Closed-source MLLMs}}} \\
Claude 3.7 Sonnet~\cite{claude} & 27.6 & 22.7 & 17.4 & 68.4 & 30.5 & 29.9 & 63.6 & 27.6 & 64.8 & 78.4 & 34.5 & 27.8 & 32.6 & 40.5 \\
Gemini 2.5 Pro~\cite{gemini25} & -- & -- & -- & -- & -- & -- & -- & -- & -- & -- & -- & -- & -- & 45.2 \\
GPT-5.2~\cite{gpt52} & 30.0 & 37.0 & 17.0 & 70.5 & 43.0 & 41.4 & \firstc \textbf{68.3} & 34.0 & \thirdc 74.0 & 76.0 & \thirdc 52.0 & 36.7 & 38.0 & 47.5 \\
\midrule
\multicolumn{15}{l}{\textit{\textcolor{gray}{Open-source MLLMs}}} \\
InternVL3-8B~\cite{internvl3} & \secondc 40.0 & 39.0 & 10.0 & 71.5 & 44.5 & 30.8 & \thirdc 65.0 & 25.2 & \firstc \textbf{77.0} & \secondc 82.0 & 36.0 & 21.7 & \thirdc 50.0 & 45.6 \\
Qwen2-VL-7B~\cite{qwen2vl} & 26.7 & 40.0 & 11.0 & 73.0 & 35.9 & 34.6 & 61.7 & 31.8 & 70.0 & \thirdc 81.0 & 35.0 & \secondc 46.7 & 48.0 & 45.8 \\
InternVL2.5-8B~\cite{internvl2} & \thirdc 38.3 & 37.0 & 10.0 & \thirdc 77.0 & 33.4 & 35.5 & \thirdc 65.0 & 21.6 & 73.0 & \firstc \textbf{83.0} & 34.0 & \firstc \textbf{50.0} & 43.0 & 46.2 \\
Qwen2.5-VL-7B~\cite{qwen25vl} & 33.3 & 40.0 & \thirdc 31.0 & \thirdc 77.0 & 40.6 & 40.5 & \secondc 66.7 & \secondc 36.2 & 68.0 & 72.0 & 27.0 & 38.3 & 45.0 & 47.4 \\
InternVL3-78B~\cite{internvl3} & 23.3 & \thirdc 49.0 & \secondc 33.0 & 74.0 & 42.5 & 37.4 & \secondc 66.7 & 30.0 & \secondc 76.0 & \thirdc 81.0 & 40.0 & \thirdc 45.0 & 42.0 & 49.2 \\
Qwen3-VL-8B~\cite{qwen3vl} & 21.7 & \secondc 50.0 & 26.0 & \firstc \textbf{81.5} & \secondc 46.6 & \secondc 43.1 & \secondc 66.7 & 30.4 & \thirdc 74.0 & 79.0 & 37.0 & 43.3 & \firstc \textbf{51.0} & 50.0 \\
DeepEyes~\cite{deepeyes} & 31.7 & 41.0 & \secondc 33.0 & 75.0 & 41.6 & 38.1 & \firstc \textbf{68.3} & 31.4 & 70.0 & 78.0 & \thirdc 46.0 & \firstc \textbf{50.0} & 46.0 & 50.0 \\
Qwen2.5-VL-72B~\cite{qwen25vl} & 33.3 & 47.0 & \secondc 39.0 & \secondc 80.0 & \thirdc 45.3 & \thirdc 42.1 & \thirdc 65.0 & 34.0 & 71.0 & 74.0 & 37.0 & \thirdc 43.3 & 42.0 & 50.2 \\
Qwen3-VL-235B-A22B~\cite{qwen3vl} & \firstc \textbf{61.7} & \firstc \textbf{82.5} & \thirdc 38.6 & 36.7 & 44.0 & 39.0 & 38.9 & \firstc \textbf{49.0} & 73.0 & \secondc 82.0 & 37.8 & 33.3 & 48.0 & 51.1 \\
Intern-S1-mini~\cite{interns1} & -- & -- & -- & -- & -- & -- & -- & -- & -- & -- & -- & -- & -- & \secondc 51.6 \\
\midrule
\textbf{GeoEyes (Our)} & \thirdc 38.3 & 40.0 & 24.0 & 73.5 & \firstc \textbf{59.5} & \firstc \textbf{66.1} & \firstc \textbf{68.3} & 32.2 & \firstc \textbf{77.0} & 80.0 & \secondc 56.0 & 40.0 & \thirdc 50.0 & \firstc \textbf{54.2} \\
\bottomrule
\end{tabular}%
}
\vspace{-6mm}
\end{table*}

\textbf{Ensuring Logical Rigor: Process Verification Reward ($R_{proc}$).}
To prevent ungrounded hallucinations, we employ a \textit{Necessity-Aware} process judge.
This module leverages the LLM's semantic priors to verify not just the consistency of the chain-of-thought, but the \textit{necessity} of tool invocation.
It penalizes the agent for generating confident answers to detail-oriented queries without performing corresponding zoom-in actions, ensuring results are derived from sufficient visual evidence.




\textbf{Optimization Strategy.} 
Finally, we optimize the policy via Group Relative Policy Optimization (GRPO).
To facilitate stable updates, we define the probability ratio $r_i(\theta) = \frac{\pi_\theta(\tau_i)}{\pi_{old}(\tau_i)}$ and maximize the following objective function:
{\setlength{\abovedisplayskip}{2pt}
\setlength{\belowdisplayskip}{2pt}
\setlength{\abovedisplayshortskip}{0pt}
\setlength{\belowdisplayshortskip}{0pt}
\setlength{\jot}{1pt} 
\vspace{-2pt}
\begin{equation}
\resizebox{0.95\columnwidth}{!}{$
\begin{aligned}
    \mathcal{J}(\theta) = \mathbb{E}_{q \sim \mathcal{D}, \tau \sim \pi_{old}} \Bigg[ & \frac{1}{G} \sum_{i=1}^{G} \min \bigg( r_i(\theta) \hat{A}_i, \\
    & \text{clip}\big(r_i(\theta), 1-\epsilon, 1+\epsilon\big) \hat{A}_i \bigg) \Bigg] - \beta D_{KL}(\pi_\theta || \pi_{ref})
\end{aligned}
$}
\end{equation}
\vspace{-20pt}}

where $q$ denotes a query sampled from the dataset $\mathcal{D}$, and $\tau$ represents a trajectory sampled from the old policy $\pi_{old}$.
The advantage $\hat{A}_i$ is computed by standardizing the cumulative rewards $R_i$ within each group of $G$ trajectories: $\hat{A}_i = (R_i - \text{mean}(\{R_j\}_{j=1}^G)) / (\text{std}(\{R_j\}_{j=1}^G) + \delta)$, where $\delta$ is a small constant for numerical stability.
Regarding hyperparameters, $\epsilon$ restricts the policy update step size, and $\beta$ controls the strength of the KL divergence penalty ($D_{KL}$) relative to the reference SFT policy $\pi_{ref}$.
This critic-free approach eliminates the need for a separate value network, reducing training instability and computational overhead while effectively navigating the sparse reward landscape of UHR scenarios.

In summary, our method initializes visual planning via cold-start SFT and utilizes AdaZoom-GRPO to adaptively optimize tool efficiency, effectively bridging the gap between mechanical invocation and intelligent adaptive focus in UHR RS scenarios.

\section{Experiments}

\subsection{Experimental Setup}
\textbf{Model, Dataset and Training.} We build our model upon the DeepEyes framework and adopt the same backbone and tool interface for all experiments to ensure fair comparison.
Different from DeepEyes, we introduce an additional supervised fine-tuning (SFT) stage before reinforcement learning. This stage is trained on our newly annotated remote sensing instruction dataset with interleaved image–text supervision and explicit tool-call annotations, providing structured initialization for tool-related behaviors.
During the reinforcement learning stage, based on the DeepEyes~\cite{deepeyes}, we further include the SuperRS-VQA dataset to enhance the diversity of remote sensing reasoning tasks.

\textbf{Evaluation.} We evaluate our model on XLRS-Bench. The MLLMs evaluated on XLRS-Bench are grouped into three categories: open-source MLLMs, closed-source MLLMs, and specialized remote sensing models. Following GeoLLaVA-8K, we report the accuracy on the L-1 dimension for the VQA task.
\subsection{Main Results}
As shown in Table~\ref{tab:vqa}, our method achieves a new state-of-the-art average accuracy of \textbf{54.23\%} on the XLRS-Bench.
Our method consistently outperforms domain-specific baselines such as GeoLLaVA-8K (51.5\%) and the agentic framework DeepEyes (50.0\%), as well as leading general-purpose MLLMs. Notably, with a 7B backbone, it surpasses much larger models including Qwen3-VL-235B (51.1\%) and Qwen2.5-VL-72B (50.2\%). The gains are most pronounced on fine-grained perception, where we achieve \textbf{66.1\%} on Object Classification (OCL) and \textbf{59.5\%} on Overall Counting (OCC), far above static global-view models (e.g., Qwen3-VL-235B at 39.0\% and 44.0\%). These results show that active zoom-in resolves the UHR resolution bottleneck without brute-force scaling, and the improvement over DeepEyes (\textbf{54.23\%} vs. 50.0\%) confirms the importance of mitigating tool-usage homogenization via SFT initialization and reward shaping.

\subsection{Ablation Study}
To isolate the sources of performance improvement over the baseline, we conduct a series of ablation experiments.
We specifically verify three core hypotheses: (1) the necessity of the domain-specific SFT cold start for initializing task-adaptive policies; (2) the superiority of our geometry-aware CoF reward over standard metrics; and (3) the impact of distinct reward components ($R_{tool}$, $R_{cof}$, $R_{proc}$) on agent behavior.  Detailed analysis are shown in the Supp.~\ref{supp:ablation}.
\begin{table}[htbp]
\centering
\scriptsize
\vspace{-2mm}
\setlength{\tabcolsep}{8.5pt}
\renewcommand{\arraystretch}{1.2}
\definecolor{headergray}{gray}{0.92}
\definecolor{rowblue}{rgb}{0.94, 0.97, 1.0} 

\begin{tabular}{l c c c}
\toprule
\rowcolor{headergray} 
\textbf{SFT Data} & \textbf{Accuracy} & \textbf{Tool Call(\%)} & \textbf{Avg. Tool Call} \\
\midrule
No SFT & 47.73 & 100 & 1.01 \\
MM-Adaptive-CoF & 47.86 & 100 & 1.14 \\

\rowcolor{rowblue} 
\textbf{UHR-CoZ (Ours)} & \textbf{52.87} & \textbf{90.36} & \textbf{1.46} \\
\bottomrule
\end{tabular}
\caption{\textbf{Ablation study on the effect of dataset in the SFT stage.}}
\label{tab:ablation_sft}
\vspace{-9mm}
\end{table}

\textbf{Effect of the SFT Cold Start.}
We first investigate the impact of the supervised fine-tuning (SFT) stage as a cold start for reinforcement learning.
Table~\ref{tab:ablation_sft} compares three initialization strategies.
This indicates that SFT provides essential initialization for on-demand tool use, which standard RL algorithms fail to learn without explicit guidance.

\textbf{Effect of CoF Reward Formulation.}  
We further analyze the impact of the reward formulation for zoom-in actions in Table~\ref{tab:cof_formulation}.
We compare our proposed \textit{Directional IoU} (Containment-based) reward against a \textit{Standard IoU} baseline.
While Standard IoU is effective for static localization, it is ill-suited for progressive zooming due to its inherent bias against significant scale changes: an effective zoom-in operation (e.g., cropping a small target from a global view) mathematically yields a low IoU score despite its correctness, causing the baseline to receive sparse rewards and learn conservative, shallow cropping policies.
To quantify this, we analyzed a \textbf{Tool-Intensive Subset} comprising four tasks with the highest demand for visual search (Average \#Tool Calls $>2$): \textit{Route Planning}, \textit{Counting with Changing Detection}, \textit{Counting with Complex Reasoning}, and \textit{Overall Counting}.
This pattern of inefficient exploration confirms that the Standard IoU metric is fundamentally ill-suited for quantifying zoom-in quality, whereas our geometry-aware reward successfully guides the agent to perform high-quality zoom-ins even under substantial scale shifts.

\begin{table}[h]
\centering
\vspace{-2mm}
\scriptsize
\renewcommand{\arraystretch}{1.3} 
\setlength{\tabcolsep}{7pt} 

\definecolor{headergray}{gray}{0.92}
\definecolor{rowblue}{rgb}{0.94, 0.97, 1.0}

\begin{tabular}{lcc}
    \toprule
    \rowcolor{headergray} 
    \textbf{\makecell[l]{CoF Reward Formulation}} & 
    \textbf{\makecell[c]{Tool-Intensive Subset (Acc.)}} & 
    \textbf{Accuracy} \\ 
    \midrule
    Standard IoU & 41.92 & 53.54 \\ 
    \rowcolor{rowblue} 
    \textbf{Ours} & \textbf{46.08} & \textbf{54.23} \\ 
    \bottomrule
\end{tabular}
\caption{\textbf{Ablation study on the CoF reward formulation.} \textit{Ours} utilizes Directional IoU. Tool-Intensive Subset reports average accuracy on the four most demanding tasks.}
\label{tab:cof_formulation}
\vspace{-7mm}
\end{table}
\textbf{Importance of Necessity-Aware Process Verification.}
We analyze the design of the Process Reward $R_{proc}$ in Table~\ref{tab:process_ablation}.
We compare a pure \textit{Logic-Consistency Judge} (Baseline) against our \textit{Necessity-Aware} formulation.
The baseline, restricted to checking internal contradictions (e.g., mismatches between intent and action), achieves 52.82\% accuracy.
However, this metric suffers from an intrinsic limitation due to the lack of visual access: it fails to penalize instances where the model confidently fabricates answers for tiny targets without necessary tool invocation, provided the reasoning remains self-consistent.
In contrast, our Necessity-Aware judge leverages semantic priors regarding task difficulty to penalize such implausible confidence.
\begin{table}[htbp]
\vspace{-2mm}
\centering
\scriptsize
\renewcommand{\arraystretch}{1.3}
\setlength{\tabcolsep}{12pt} 

\definecolor{headergray}{gray}{0.92}
\definecolor{rowblue}{rgb}{0.94, 0.97, 1.0}

\begin{tabular}{lc}
    \toprule
    \rowcolor{headergray} 
    \textbf{Process Reward Variant} & \textbf{Accuracy} \\
    \midrule
    Logic-Consistency & 52.82 \\
    \rowcolor{rowblue} 
    \textbf{Necessity-Aware (Ours)} & \textbf{54.23} \\
    \bottomrule
\end{tabular}
\caption{\textbf{Ablation study on the process reward.} The \textit{Logic-Consistency Judge} only evaluates reasoning-answer consistency. The \textit{Necessity-Aware Judge} additionally penalizes unnecessary actions.}
\label{tab:process_ablation}
\vspace{-9mm}
\end{table}

\textbf{Impact of Reward Components.}
We perform a leave-one-out analysis in Table~\ref{tab:reward_ablation} to isolate the contribution of each reward term.
Removing $R_{tool}$ results in a consistent performance decline across nearly all task categories, leading to an approximate 2\% drop in macro accuracy.
Furthermore, excluding the \textit{Chain-of-Focus Reward} ($R_{cof}$) yields the lowest accuracy (51.56\%), confirming that efficiency constraints without geometric guidance trap the agent in ineffective zooming policies.
Similarly, omitting the \textit{Process Verification Reward} ($R_{proc}$) also leads to a performance decline (53.05\%), verifying its contribution to the final results.

\begin{table}[htbp]
\vspace{-2mm}
\centering
\scriptsize
\renewcommand{\arraystretch}{1.3}
\setlength{\tabcolsep}{12pt}
\setlength{\aboverulesep}{0pt}
\setlength{\belowrulesep}{0pt}
\definecolor{headergray}{gray}{0.92}
\definecolor{rowblue}{rgb}{0.94, 0.97, 1.0}
\begin{tabular}{cccc}
\toprule
\rowcolor{headergray} \multicolumn{3}{c}{\textbf{Reward Components}} & \\
\rowcolor{headergray} \cmidrule(lr){1-3}
\rowcolor{headergray} \textbf{$R_{\text{tool}}$} & \textbf{$R_{\text{cof}}$} & \textbf{$R_{\text{proc}}$} & \multirow{-1.8}{*}{\textbf{Accuracy}} \\
\specialrule{\lightrulewidth}{0pt}{0pt}
\rowcolor{rowblue} \checkmark & \checkmark & \checkmark & \textbf{54.23} \\
\specialrule{\lightrulewidth}{0pt}{0pt}
\texttimes & \checkmark & \checkmark & 52.24 \\
\checkmark & \texttimes & \checkmark & 51.56 \\
\checkmark & \checkmark & \texttimes & 53.05 \\
\bottomrule
\end{tabular}
\caption{\textbf{Impact of different reward components during reinforcement learning.} The first row denotes our full method. $R_{\text{cof}}$, $R_{\text{proc}}$, and $R_{\text{tool}}$ refer to the Chain-of-Focus, Process Verification, and Adaptive Efficiency rewards, respectively.}
\label{tab:reward_ablation}
\vspace{-9mm}
\end{table}


\section{Conclusion}

In this paper, we diagnose a fundamental bottleneck for zoom-enabled MLLMs in UHR RS: Tool Usage Homogenization, where tool policies collapse into saturated, task-agnostic, near-one-call behavior that is inherently mismatched to task heterogeneity and low effective evidence density, and therefore cannot reliably acquire the right evidence at the right depth. 
We address this failure mode with a staged recipe that couples process-supervised initialization with evidence-driven RL shaping: UHR-CoZ provides interleaved demonstrations spanning no-tool, single-zoom, and multi-round progressive focusing, while AdaZoom-GRPO reshapes optimization toward evidence gain and answer improvement via adaptive efficiency, geometry-aware progressive zooming, and necessity-aware process verification.
The resulting model, GeoEyes, achieves 54.23\% average accuracy on XLRS-Bench, improving over both domain-specialized baselines (GeoLLaVA-8K 51.5\%, DeepEyes 50.0\%) and even substantially larger general-purpose models (Qwen3-VL-235B 51.1\%) despite using a 7B backbone, with the strongest gains on fine-grained perception (66.1\% OCL, 59.5\% OCC), demonstrating that active, policy-controlled zooming resolves the UHR resolution bottleneck without brute-force scaling. 
Overall, our results suggest that robust UHR RS VQA requires explicitly training tool policies to differentiate when to abstain, when to iterate, and when to stop, and that combining domain-aligned process supervision with evidence-centric reward shaping is a principled path to achieving this.

\section*{Impact Statement}

This paper presents work whose goal is to advance the field of Machine Learning. There are many potential societal consequences of our work, none which we feel must be specifically highlighted here.

\bibliography{example_paper}
\bibliographystyle{bibstyle2026}

\newpage
\appendix
\onecolumn

\begin{figure*}[t]
  \centering
  \includegraphics[width=\textwidth]{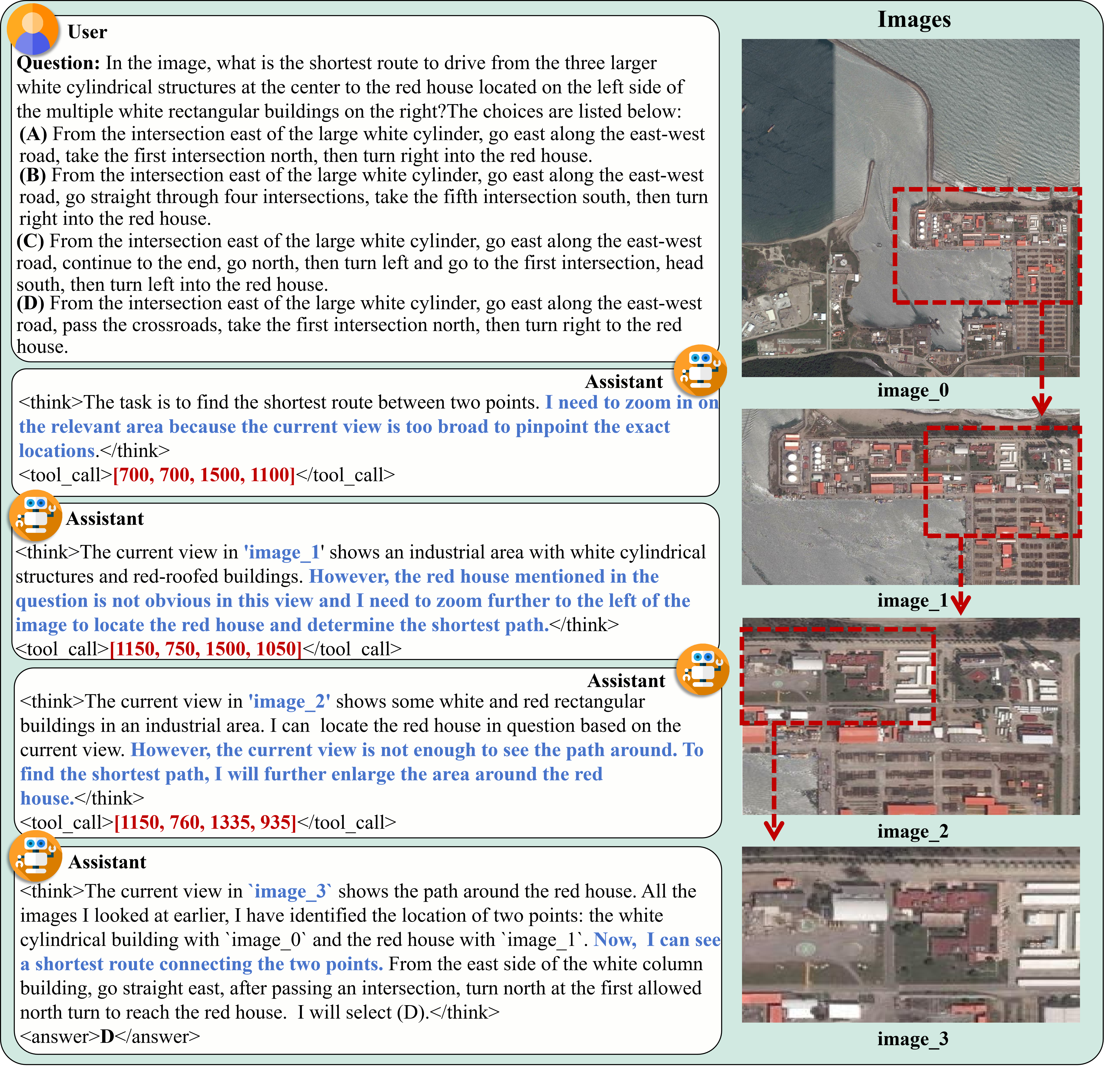}
\caption{\textbf{UHR-CoZ example with three zoom-in calls.} Left: the interleaved dialogue, where the model alternates between stepwise reasoning and \texttt{tool\_call} requests that specify normalized bounding boxes. Right: the corresponding multi-scale views returned by the agent, from the global image (\texttt{image\_0}) to three progressively localized crops (\texttt{image\_1}--\texttt{image\_3}). Red dashed boxes and arrows indicate the selected regions across rounds, illustrating iterative evidence acquisition for answering the question.}
  \label{fig:uhrcof_3zoomin_example}
\end{figure*}

\section{Data Construction Pipeline}
\label{app:uhrcof_examples}
\noindent \textbf{Detailed Annotation Generation for UHR-CoZ.} 
We address this challenge with a agent pipeline, where an agent orchestrates the model and a \texttt{zoom\_in} tool to generate end-to-end reasoning trajectories. 
To this end, we employ an agent-based system to generate multi-round focusing annotations. The agent orchestrates the model and a zoom-in tool to produce end-to-end trajectories, as illustrated in Figure~\ref{fig:data_pipeline}. For each sample, the input consists only of the question and the original remote sensing image (\texttt{image\_0}) with its metadata. At each round, the model reasons over the currently visible context and chooses between two actions: if it deems the evidence sufficient, it outputs the final answer and terminates; otherwise, it issues a \texttt{zoom\_in} request in a fixed format specifying the source image, a normalized bounding box, and a brief rationale. The agent executes the zoom-in operation, returns the cropped and magnified local view (indexed sequentially as \texttt{image\_1}, \texttt{image\_2}, etc.), and appends the new image together with the bounding-box information to the dialogue history, enabling subsequent rounds to leverage both global and local evidence. This loop continues until the model answers or reaches a preset maximum number of rounds. To ensure usability and consistency, the agent enforces strict per-round validation during generation: (i) format checks to ensure required fields are present and coordinates fall within valid ranges; (ii) execution checks to verify that cropping is feasible, the source image is traceable within the dialogue history, and a valid local view is returned; and (iii) interaction constraints requiring each zoom-in to either further refine an existing local view or switch to a new candidate region from a higher-level view for correction, while avoiding redundant exploration of the same area.

To illustrate the decision process in multi-round interactions, we use a two-zoom trajectory as an example. In the first round, the model determines from \texttt{image\_0} that the evidence is insufficient and issues a \texttt{zoom\_in} request by proposing a bounding box that covers a candidate region on the full image; the agent returns the corresponding local view \texttt{image\_1}. In the second round, the model verifies the evidence using both \texttt{image\_0} and \texttt{image\_1}: if the region is on track but still lacks detail, it proposes a smaller bounding box within \texttt{image\_1} to further magnify the area and obtain \texttt{image\_2}; if the initial localization deviates from the target, it can return to \texttt{image\_0} and select a new candidate box that does not overlap previously explored regions. In the third round, the model makes the final decision and outputs the answer conditioned on the multi-scale views (\texttt{image\_0}, \texttt{image\_1}, \texttt{image\_2}) and the full dialogue history. The same procedure applies to trajectories with three or more zoom-ins: the model repeatedly localizes and refines over the accumulated image set until it gathers sufficient local evidence or reaches the maximum round limit. This constrained multi-round interaction yields not only the final answer but also a traceable, multi-scale evidence acquisition process and tool-call trajectory, providing explicit process supervision for SFT cold start and encouraging differentiated tool-use policies across question types. 

\section{Details analysis of Ablation Study}
\label{supp:ablation}

\textbf{Impact of Reward Components.}
We perform a leave-one-out analysis in Table~\ref{tab:reward_ablation} to isolate the contribution of each reward term.
Removing $R_{tool}$ results in a consistent performance decline across nearly all task categories, leading to an approximate 2\% drop in macro accuracy.
Crucially, this degradation occurs despite the tool call ratio remaining nearly identical (68.44\% vs. 67.47\%).
This observation indicates that the efficiency reward improves the precision of tool invocation rather than simply altering its frequency.
Without this constraint, the agent fails to distinguish between necessary and redundant actions, leading to ineffective exploration and significant drops in complex tasks such as \textit{Overall Counting} (30.00\% vs. 38.33\%) and \textit{Route Planning} (52.00\% vs. 56.00\%).

\begin{table*}[htbp]
\footnotesize
\centering
\caption{\textbf{Ablation study results on the perception and reasoning dimensions on XLRS-Bench.} `Avg.' represents the macro average accuracy across sub-tasks.}
\label{tab:ablation_add}
\vspace{0.2cm}
\resizebox{\textwidth}{!}{%
\begin{tabular}{l|cccccccc|ccccc|c}
\toprule 
\rowcolor[gray]{.9} \textbf{Method} & \multicolumn{8}{c|}{\textbf{Perception}} & \multicolumn{5}{c|}{\textbf{Reasoning}} & \textbf{Avg.} \\ 
\midrule
\rowcolor[gray]{.9} \textbf{Sub-tasks (L-3 Capability)} & \textbf{OC} & \textbf{RC} & \textbf{OLUC} & \textbf{RLUC} & \textbf{OCC} & \textbf{OCL} & \textbf{OMS} & \textbf{OSR} & \textbf{AD} & \textbf{ECR} & \textbf{RP} & \textbf{RCCD} & \textbf{CCR} & \\ 
\midrule
\multicolumn{15}{l}{\textit{\textcolor{gray}{Ablation of SFT Data}}} \\

No SFT & 16.67 & 44.00 & 21.00 & 76.50 & 45.62 & 39.25 & 66.67 & 32.80 & 73.00 & 75.00 & 37.00 & 45.00 & 48.00 & 47.73 \\
MM-Adaptive-CoF & 16.67 & 32.00 & 23.00 & 77.50 & 54.75 & 41.50 & 66.67 & 29.80 & 66.00 & 78.00 & 43.00 & 48.33 & 45.00 & 47.86 \\
Our data & 18.33 & 43.00 & 48.0 & 79.50 & 46.62 & 56.50 & 65.00 & 30.00 & 72.00 & 86.00 & 49.00 & 43.33 & 50.00 & 52.87 \\

\midrule
\multicolumn{15}{l}{\textit{\textcolor{gray}{Ablation of CoF Formulation and LLM Judge}}} \\ 
Ours & 38.33 & 40.00 & 24.00 & 73.50 & 59.50 & 66.13 & 68.33 & 32.20 & 77.00 & 80.00 & 56.00 & 40.00 & 50.00 & 54.23 \\
Standard IoU & 33.33 & 43.00 & 25.00 & 73.00 & 61.12 & 66.63 & 70.00 & 33.60 & 74.00 & 82.00 & 49.00 & 38.33 & 47.00 & 53.54 \\
Logic-Consistency & 35.00 & 42.00 & 19.00 & 75.50 & 60.62 & 66.25 & 61.67 & 34.60 & 75.00 & 78.00 & 52.00 & 35.00 & 53.00 & 52.82 \\

\bottomrule
\end{tabular}%
}
\end{table*}

\begin{table*}[htbp]
\footnotesize
\centering
\caption{\textbf{Ablation study results on the perception and reasoning dimensions on XLRS-Bench.} `Avg.' represents the number of tool calls averaged on the whole dataset.}
\label{tab:ablation_add2}
\vspace{0.2cm}
\resizebox{\textwidth}{!}{%
\begin{tabular}{l|cccccccc|ccccc|c}
\toprule 
\rowcolor[gray]{.9} \textbf{Method} & \multicolumn{8}{c|}{\textbf{Perception}} & \multicolumn{5}{c|}{\textbf{Reasoning}} & \textbf{Avg.} \\ 
\midrule
\rowcolor[gray]{.9} \textbf{Sub-tasks (L-3 Capability)} & \textbf{OC} & \textbf{RC} & \textbf{OLUC} & \textbf{RLUC} & \textbf{OCC} & \textbf{OCL} & \textbf{OMS} & \textbf{OSR} & \textbf{AD} & \textbf{ECR} & \textbf{RP} & \textbf{RCCD} & \textbf{CCR} & \\ 
\midrule
\multicolumn{15}{l}{\textit{\textcolor{gray}{Ablation of SFT Data}}} \\

No SFT & 1.05 & 1.00 & 1.11 & 1.00 & 1.00 & 1.00 & 1.00 & 1.00 & 1.00 & 1.01 & 1.00 & 1.47 & 1.00 & 1.01 \\
MM-Adaptive-CoF & 1.15 & 1.01 & 1.69 & 1.00 & 1.01 & 1.00 & 1.25 & 1.25 & 1.04 & 1.11 & 2.05 & 1.88 & 1.07 & 1.14 \\
Our data & 3.32 & 1.75 & 0.33 & 0.90 & 0.92 & 1.00 & 0.93 & 2.81 & 1.05 & 0.88 & 2.65 & 2.75 & 2.84 & 1.46 \\

\midrule
\multicolumn{15}{l}{\textit{\textcolor{gray}{Ablation of CoF Formulation and LLM Judge}}} \\ 
Ours & 2.3 & 1.24 & 0.08 & 0.33 & 0.58 & 0.99 & 0.73 & 1.86 & 0.43 & 0.47 & 2.08 & 1.42 & 2.69 & 1.05 \\
Standard IoU & 2.40 & 1.40 & 0.13 & 0.30 & 0.60 & 0.99 & 0.73 & 1.97 & 0.40 & 0.61 & 2.26 & 2.52 & 2.13 & 1.09 \\
Logic-Consistency & 2.25 & 1.37 & 0.12 & 0.38 & 0.62 & 0.99 & 0.70 & 1.97 & 0.52 & 0.51 & 2.18 & 2.48 & 2.89 & 1.11 \\



\bottomrule
\end{tabular}%
}
\end{table*}

\begin{table*}[htbp]
\renewcommand{\arraystretch}{1.5}
\footnotesize
\caption{\textbf{Characteristics of XLRS-Bench} The full names of the Level-3 task abbreviations are also provided.}
\centering
\resizebox{0.98\textwidth}{!}{
\begin{tabular}{c|c|cccccc}
\toprule  \Gray
\textbf{L1-Task }& \textbf{L2-Task} &\textbf{ L3-Task }& \textbf{Abbr.} & \textbf{Annotation Format} & \textbf{Number of Samples} & \textbf{Answer Type}\\ \hline
\multirow{8}{*}{Perception} & \multirow{2}{*}{Counting} & Overall Counting & OC & VQA & 60 & Multiple Choice(A/B/C/D)\\
 & & Regional Counting & RC & VQA & 100 & Multiple Choice(A/B/C/D)\\
\cline{2-7}
 & \multirow{2}{*}{Scene Classification} & Overall Land Use Classification & OLUC & VQA & 100 & Multiple Choice(A/B/C/D)\\
 & & Regional Land Use Classification & RLUC & VQA & 200 & Multiple Choice(A/B/C/D)\\
\cline{2-7}
 & Object Spatial Relationship & Object Spatial Relationship & OSR & VQA & 500 & Multiple Choice(A/B/C/D)\\
\cline{2-7}
 & \multirow{3}{*}{Object Properties} & Object Classification & OCC & VQA & 800 & Multiple Choice(A/B/C/D)\\
 & & Object Color & OCL & VQA & 800 & Multiple Choice(A/B/C/D)\\
 & & Object Motion State & OMS & VQA & 60 & Multiple Choice(A/B for Yes/No)\\
\hline
\multirow{5}{*}{Reasoning} & Route Planning & Route Planning & RP & VQA & 100 & Multiple Choice(A/B/C/D)\\
\cline{2-7}
 & Anomaly Reasoning & Anomaly Detection and Interpretation & AD & VQA & 100 & Multiple Choice(A/B/C/D)\\
\cline{2-7}
 & \multirow{2}{*}{Complex Reasoning} & Environmental Condition Reasoning & ECR & VQA & 100 & Multiple Choice(A/B/C/D)\\
 & & Counting with Complex Reasoning & CCR & VQA & 100 & Multiple Choice(A/B/C/D)\\
\cline{2-7}
 & Spatiotemporal Reasoning & Regional Counting with Change Detection & RCCD & VQA & 60 & Multiple Choice(A/B/C/D)\\
\bottomrule
\end{tabular}
}
\label{tab:statistics}
\vspace{-2mm}
\end{table*}

\clearpage
\begin{tcolorbox}[
    breakable, 
    enhanced,
    title=Prompt for Iterative Zoom-in Data Generation, 
    fonttitle=\bfseries,
    colback=white,
    colframe=gray!75!black
]
\textbf{\# 1. Your Role and Goal} \\
You are an expert Remote Sensing Image Analyst. Your mission is to analyze satellite and aerial imagery to answer specific questions with high accuracy and efficiency. You will reason internally and output your chosen action as a structured JSON object.

\vspace{0.5em}
\textbf{\# 2. Your Core Task and Output Format} \\
Your operation MUST follow a strict "internal reasoning -> external action" process.
\begin{itemize}
    \item \textbf{Internal Reasoning:} You will first reason internally about the problem to decide on your next action. This reasoning process will be captured separately.
    \item \textbf{External Action (Your Direct Output):} After reasoning, your direct output for each turn MUST be a single, valid JSON object that represents your chosen action. This JSON object MUST be enclosed between \texttt{<|begin\_of\_box|>} and \texttt{<|end\_of\_box|>}. You MUST NOT output any other text, tags, or explanations.
\end{itemize}

\vspace{0.5em}
\textbf{\# 3. Your Tools (How You Act)} \\
- You have one tool available: \texttt{zoom\_in}. \\
- \textbf{Bounding Box Format:} All bounding box coordinates in your \texttt{tool\_call} MUST be normalized to a 1000x1000 coordinate system (\texttt{[x\_min, y\_min, x\_max, y\_max]}), where [0, 0] is the top-left corner and [1000, 1000] is the bottom-right corner. \\
- \textbf{Tool Definition:} \\
\begin{verbatim}
<tools>
{
  "type": "function",
  "function": {
    "name": "zoom_in",
    "description": "Crops a high-resolution version of a region...",
    "parameters": {
      "type": "object",
      "properties": {
        "source_image_id": {"type": "string"},
        "reason": {"type": "string"},
        "bbox": {
          "type": "array",
          "items": {"type": "number"},
          "minItems": 4, "maxItems": 4
        }
      },
      "required": ["source_image_id", "reason", "bbox"]
    }
  }
}
</tools>
\end{verbatim}

\vspace{0.5em}
\textbf{\# 4. Action Schemas (The JSON you MUST generate)} \\
\textbf{If you decide to call a tool:}
\begin{verbatim}
{
  "type": "tool_call",
  "tool_name": "zoom_in",
  "arguments": {
    "source_image_id": "<image_id>",
    "reason": "<justification>",
    "bbox": [x_min, y_min, x_max, y_max]
  }
}
\end{verbatim}
\textbf{If you decide to provide the final answer:}
\begin{verbatim}
{
  "type": "answer",
  "content": "<Your final answer text>"
}
\end{verbatim}

\vspace{0.5em}
\textbf{\# 5. Reasoning and Strategy Guidelines} \\
- \textbf{Efficiency First:} Only generate a \texttt{tool\_call} if it's absolutely necessary. \\
- \textbf{Iterative Refinement:} If a zoom is insufficient, perform a more focused zoom on the new image. \\
- \textbf{Meaningful Zoom:} Each \texttt{zoom\_in} must target a smaller sub-region. \\
- \textbf{Plausible Aspect Ratio:} Avoid extreme "slivers".

\vspace{0.5em}
\textbf{\# 6. Few-Shot Examples (In-Context Learning)} \\
\textit{[START OF SIMPLE EXAMPLE]} \\
\textbf{USER:} [IMAGE\_0] \texttt{<metadata>\{"image\_id": "image\_0", ...\}</metadata>} \\
Question: "What is the tail number of the large passenger aircraft parked at the terminal?" \\
\textbf{ASSISTANT:} \\
\texttt{<think>}The tail number is illegible in image\_0. A single zoom on the tail is needed.\texttt{</think>} \\
\texttt{<|begin\_of\_box|>} \\
\{"type": "tool\_call", "tool\_name": "zoom\_in", "arguments": \{"source\_image\_id": "image\_0", "reason": "To read the tail number.", "bbox": [710, 450, 780, 500]\}\} \\
\texttt{<|end\_of\_box|>} \\
\textbf{TOOL\_RESPONSE:} [IMAGE\_1] \texttt{<metadata>... "origin\_context": \{"source\_image\_id": "image\_0", ...\} </metadata>} \\
\textbf{USER:} Here is the new visual information... \\
\textbf{ASSISTANT:} \\
\texttt{<think>}The close-up in image\_1 clearly shows the tail number.\texttt{</think>} \\
\texttt{<|begin\_of\_box|>} \{"type": "answer", "content": "N780AN"\} \texttt{<|end\_of\_box|>} \\
\textit{[END OF SIMPLE EXAMPLE]}

\vspace{0.5em}
\textit{[START OF COMPLEX EXAMPLE]} \\
\textbf{USER:} [IMAGE\_0] Question: "What is the type of the bridge and is there recent construction on its northern approach?" \\
\textbf{ASSISTANT:} \texttt{<think>}Need to identify bridge type and check construction. Zooming on bridge first.\texttt{</think>} \\
\texttt{<|begin\_of\_box|>} \{"type": "tool\_call", "tool\_name": "zoom\_in", "arguments": \{...\}\} \texttt{<|end\_of\_box|>} \\
\textit{(... iterations continue ...)} \\
\textbf{ASSISTANT:} \texttt{<think>}Image\_2 shows construction vehicles. Both parts answered.\texttt{</think>} \\
\texttt{<|begin\_of\_box|>} \{"type": "answer", "content": "The bridge is a suspension bridge, and there is clear evidence of recent construction..."\} \texttt{<|end\_of\_box|>} \\
\textit{[END OF COMPLEX EXAMPLE]}
\end{tcolorbox}
\captionof{table}{System prompt and In-Context Learning (ICL) examples used for generating multi-turn zoom-in reasoning trajectories in UHR-CoZ.}
\label{tab:app_zoom_in_prompt}

\end{document}